\documentclass[letterpaper]{article} 
\usepackage{aaai2026}  
\usepackage{times}  
\usepackage{helvet}  
\usepackage{courier}  
\usepackage[hyphens]{url}  
\usepackage[most]{tcolorbox} 
\usepackage{graphicx} 
\usepackage{xcolor} 
\usepackage{booktabs} 
\usepackage{natbib}  
\usepackage{caption} 
\tcbset{
  myblurbA/.style={
    colback=blue!5,   
    colframe=blue!40, 
    coltitle=black,
    fonttitle=\bfseries,
    boxrule=0.6pt,
    arc=2mm,
    left=4mm, right=4mm, top=2mm, bottom=2mm
  },
  myblurbB/.style={
    colback=red!5,    
    colframe=red!40,  
    coltitle=black,
    fonttitle=\bfseries,
    boxrule=0.6pt,
    arc=2mm,
    left=4mm, right=4mm, top=2mm, bottom=2mm
  }
}
\usepackage{algorithm}
\usepackage{algorithmic}
\usepackage{color}
\usepackage{amsmath}
\usepackage{amsfonts} 
\usepackage{graphicx, subcaption}
\usepackage{newfloat}
\usepackage{listings}
\DeclareCaptionStyle{ruled}{labelfont=normalfont,labelsep=colon,strut=off} 
\lstdefinestyle{yamlstyle}{
    language=Python,
    basicstyle=\rmfamily\small,  
    keywordstyle=\color{blue}\bfseries,
    commentstyle=\color{green!60!black},
    stringstyle=\color{red!70!black},
    backgroundcolor=\color{red!5!white},  
    frame=single,
    frameround=tttt,
    breaklines=true,
    showstringspaces=false,
    tabsize=2,
    xleftmargin=5pt,
    xrightmargin=5pt,
    aboveskip=10pt,
    belowskip=10pt,
    lineskip=1pt
}
\floatstyle{ruled}
\newfloat{listing}{tb}{lst}{}
\floatname{listing}{Listing}
\title{Large Language Model-Based Reward Design for Deep Reinforcement Learning-Driven Autonomous Cyber Defense}
\author{
    Sayak Mukherjee\textsuperscript{\rm 1},
    Samrat Chatterjee\textsuperscript{\rm 1},     
    Emilie Purvine\textsuperscript{\rm 1}, 
    Ted Fujimoto\textsuperscript{\rm 1}, 
    Tegan Emerson\textsuperscript{\rm 1}
}
\affiliations{
    \textsuperscript{\rm 1}Pacific Northwest National Laboratory\\

    902 Battelle Boulevard\\
    Richland, Washington 99354, USA\\
    sayak.mukherjee@pnnl.gov, samrat.chatterjee@pnnl.gov
}

\usepackage{listings}
\usepackage{xcolor}

\lstdefinestyle{yamlstyle}{
    language=Python,
    basicstyle=\ttfamily\footnotesize,
    keywordstyle=\color{blue}\bfseries,
    commentstyle=\color{green!60!black},
    stringstyle=\color{red},
    backgroundcolor=\color{gray!10},
    frame=single,
    breaklines=true,
    showstringspaces=false,
    tabsize=2,
    captionpos=b
}

\usepackage{bibentry}

\begin{document}

\maketitle

\begin{abstract}
Designing rewards for autonomous cyber attack and defense learning agents in a complex, dynamic environment is a challenging task for subject matter experts. We propose a large language model (LLM)-based reward design approach to generate autonomous cyber defense policies in a deep reinforcement learning (DRL)-driven experimental simulation environment. Multiple attack and defense agent personas were crafted, reflecting heterogeneity in agent actions, to generate LLM-guided reward designs where the LLM was first provided with contextual cyber simulation environment information. These reward structures were then utilized within a DRL-driven attack-defense simulation environment to learn an ensemble of cyber defense policies. Our results suggest that LLM-guided reward designs can lead to effective defense strategies against diverse adversarial behaviors.

\end{abstract}

\section{Introduction}
\label{sec:introduction}
With the ease of access to emerging artificial intelligence (AI) tools, such as large language models (LLMs) and agentic AI frameworks that can automate coding workflows and decisions, cyber defenders with resource constraints are tasked with the growing challenge of protecting critical systems and networks against increasingly sophisticated attackers. The recently released America's AI Action Plan~\cite{ai2025plan}, emphasizes the importance of secure-by-design, robust, and resilient AI-enabled cyber defense tools to safeguard critical infrastructure. Aligned with emerging AI-enabled cybersecurity priorities, there is a growing body of work in application of deep reinforcement learning (DRL) and LLMs for cybersecurity with focus on determining optimal cyber defense strategies against different threat types~\cite{nguyen2019deep, dutta2021constraints, dutta2023deep, sewak2023deep, oesch2024path, oesch2024towards, zhang2025llms, xu2025l2m, castro2025large, loevenich2025design}. Continued focus on rigorous experimental evaluation of DRL and LLM approaches while leveraging high fidelity simulation environments to train automated agents under dynamic conditions is a key step toward autonomous cyber defense. Within the DRL for cybersecurity context, designing reward structures for diverse adversarial and defense agents is challenging and requires balance between immediate and long-term agent goals to achieve effective learned policies. Typically, cyber operations subject matter experts (SMEs) are sought to shape attack and defense agent rewards. However, with increasing cyber system state space and complexity of agent actions, designing rewards that adequately reflect agent heterogeneity can lead to significant cognitive burden for SMEs.  

Meanwhile, with the growing prevalence of LLMs, there has also been focus on how to use such models to augment RL agents, with problem taxonomy defined in~\cite{pternea2024rl} classifying the research areas into: 1) RL4LLM: These studies use RL to improve the performance of an LLM in a natural language processing task, 2) LLM4RL: These studies use an LLM to supplement the training of an RL model that performs a general task that is not inherently related to natural language, and 3) RL+LLM: These studies combine RL models with LLM models to plan over a set of skills, without using either model to train or fine-tune the other.
We focus on LLM4RL methods to design reward structures. Efficiently designing reward functions that accurately represent the task requires LLMs that have state-of-the-art coding abilities. We utilize the Claude Sonnet model that are instruction-tuned LLMs with coding and reasoning abilities \cite{anthropic2025claude4}. \cite{ma2024eureka} recently also demonstrated human-level reward design via coding LLMs.

This paper focuses on the applicability of LLMs to construct context-aware reward structure designs for attack and defense agents based on their heterogeneous behavioral characteristics, and subsequently develop DRL-based autonomous cyber defense policies. We utilize a state-of-the-art coding LLM, Claude Sonnet, to generate reward structures with context about a cyber simulation environment (e.g., Cyberwheel~\cite{oesch2024path, oesch2024towards}) and baseline cyber SME-provided reward structures. These reward structures are then utilized to train DRL-based autonomous defense mechanisms for various combinations of attack and defense agents with distinct personas (e.g., an aggressive or stealthy attacker against a proactive defender). We conducted cyber simulation experiments to systematically evaluate the role of LLMs in designing diverse agent (attack and defense) rewards and compare the performance of learned cyber defense agent policies. The rest of the paper is organized as follows. The next section describes the proposed LLM-assisted autonomous DRL-driven cyber defense framework, along with a description of the simulation environment. This is followed by detailed experimental results and discussion. We conclude with main takeaways and future research steps.

\color{black}

\section{Methodology}
\label{sec:framework}
\begin{figure*}[t!]
    \centering
    \includegraphics[width=0.9\textwidth]{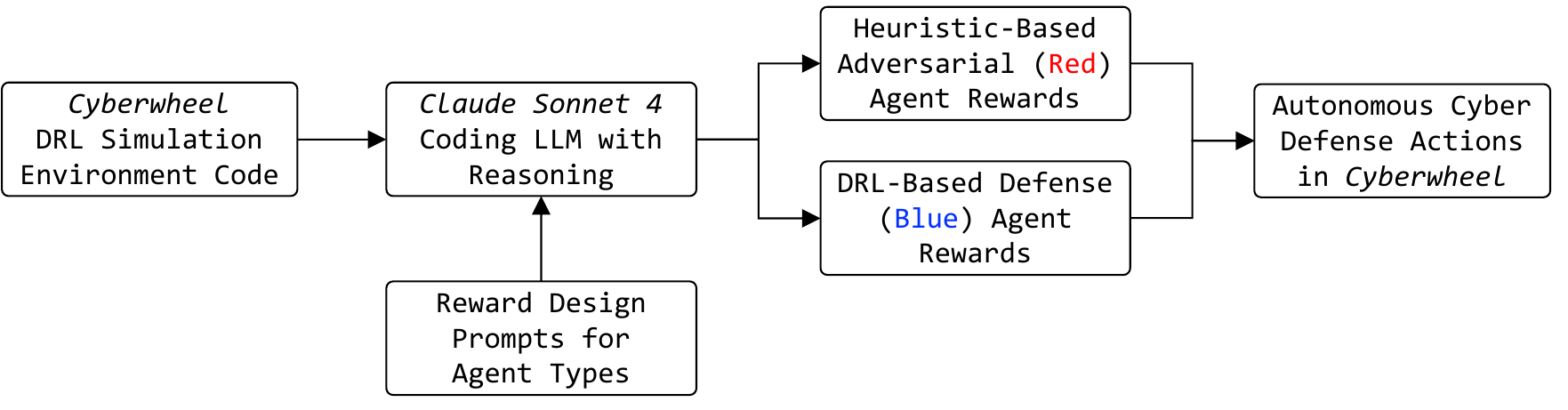}
    \caption{Methodological overview with LLM for reward design in a cyber DRL simulation environment.}
    \label{fig:overview}
\end{figure*}

The autonomous cyber defense problem focuses on learning the optimal policies for the defender (blue) agent against different forms of adversarial actions taken by the attacker (red) agent. This sequential decision-making problem can be mathematically represented with a Markov decision process (MDP). Fig. \ref{fig:overview} presents the overview scheme of the LLM-assisted reward designs for the red and blue agents to perform the DRL training of autonomous cyber defense actions. In our setup, the adversarial agent uses a standard form of attack/kill-chain sequences as prescribed by the Atomic Red Team (ART) and MITRE ATT\&CK framework. For the blue agent's policy design, let us denote the state space by $S$, defender's action space by $A$, rewards by $r$, and $\gamma$ is the discount factor. The MDP is fully characterized by the tuple $(S,A,r,\gamma)$. The optimal policy for this SDP model is a sequence of optimal actions, $a^*\in A$, characterized by the actions sampled from the optimal policy $\pi_{\theta^\star}(s)$ for any current state, $s\in S$, that can maximize the discounted finite horizon cumulative reward. In the DRL designs, the actions are sampled from a learned parametric policy $\pi_\theta(s)$, where the deep neural network parameters are denoted by $\theta$. In this context, the state space is interpreted similarly to the observation space, as we only have observables for the cyber operations. The reward design encompasses different forms of agent types, both adversarial and defense agents, and utilizes a joint reward structure of both agents prescribed by an LLM during training. We will describe how the LLM guides the designer in providing the reward structures for the policy training in subsequent sub-sections. In this work, the RL agent's policy design is aided by an LLM, where at time step $t$, the agent executes an action $a_t$ at current state $s_t$ and receives a reward $r_t$ that is used to update the policy. The LLM agentic system provides the reward structures that correspond to the gains and penalties corresponding to different actions taken by the blue (defense) and red (adversarial) agents, i.e., $\mathcal{R}_{\text{LLM}} = \mathcal{L}_{\text{LLM}}(\text{task prompt, context, behavioral constraints})$, where $\mathcal{L}_{\text{LLM}}(\cdot)$ represents a functional mapping that converts the designer prompts to the desired reward structure that can capture behavioral characterizations of red and blue agents, and then generate the reward $r_t = f(\mathcal{R}_{\text{LLM}})$, where $f(\cdot)$ will be described shortly. 

\subsection{Cyberwheel Simulation Environment}

We perform our design using the Cyberwheel simulation environment \cite{oesch2024towards}. Cyberwheel is a high-fidelity simulation environment to study autonomous cyber defense mechanisms. The simulation environment is fast, scalable, and flexible to customize rewards, actions, and observations. The customizable cyber environment is built with \texttt{networkx} graphs, and comprises hosts, subnets, and routers. These components share a common routing table, default routes, and a firewall. Each connected subnet has a corresponding router interface, enabling communication across the network. Subnets keep track of the connected hosts and corresponding IPs. Hosts can be of different types, such as mail server, file server, or user workstations. 

\begin{figure}[t!]
    \centering
    \includegraphics[width = \linewidth]{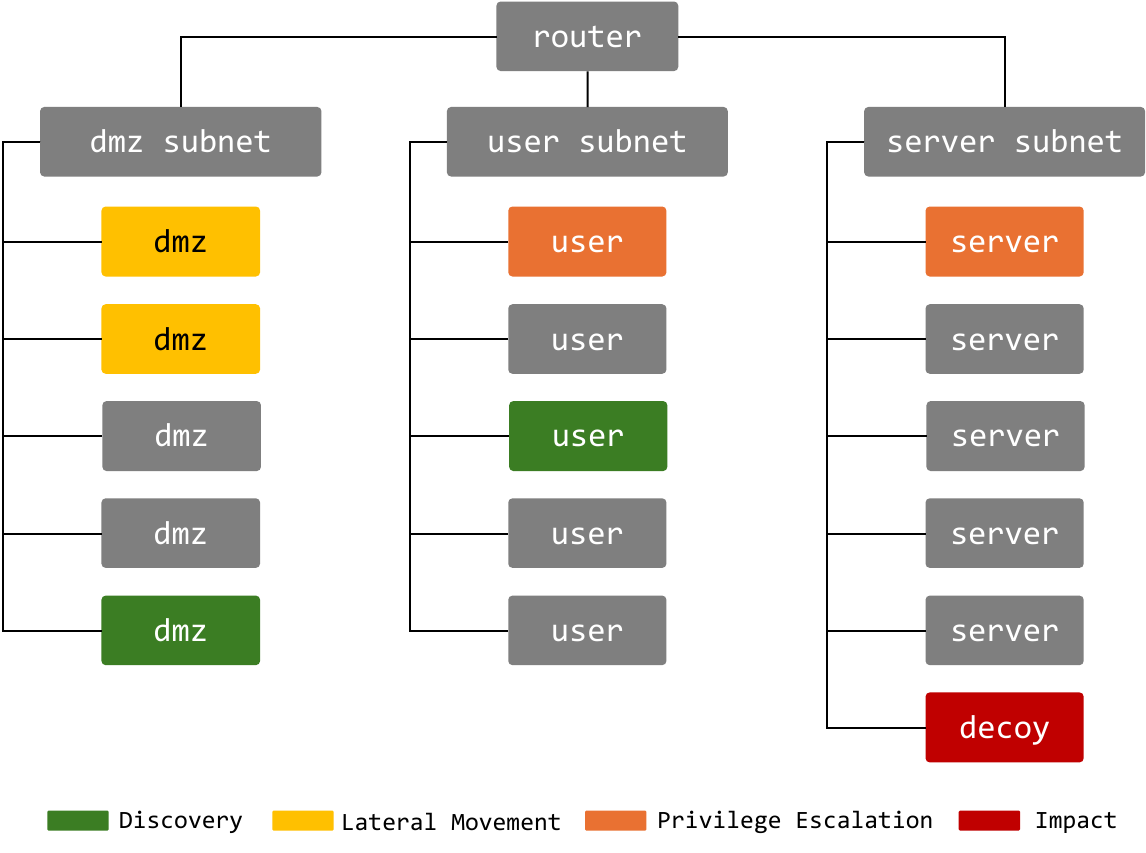}
    \caption{Example state space in a \textit{Cyberwheel} 15-host network with one router, three subnets, and multiple hosts.}
    \label{fig:network_example}
\end{figure}

The network can be configured in a customized way using a \texttt{yaml} file, where the details of the routers, subnets, and hosts are provided. Fig. \ref{fig:network_example} shows an example of the network topology that contains one router, three subnets, and five hosts, which has also been used in our numerical experimentation. The environment defines the defender (blue) and attacker (red) agents, along with their individual specified actions, rewards, and observation \texttt{yaml} files. We will describe the characteristics of these agents in detail in the subsequent sub-sections.

\subsection{Adversarial and Defense Mechanism}

\begin{figure}[h!]
\centering
\includegraphics[width = 0.85\linewidth]{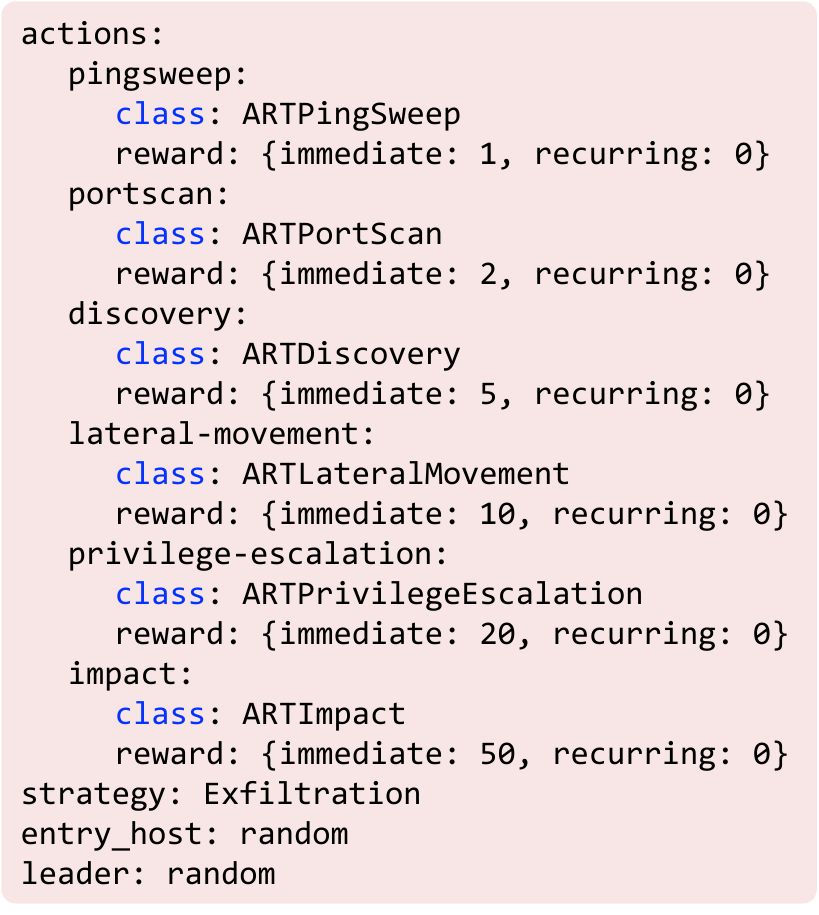}
\caption{Baseline red agent \texttt{yaml} file with actions and rewards.}
\label{fig:red_agent_yaml}
\end{figure}

The adversary, red agent, has been implemented in a logic-driven killchain-based workflow following from Atomic Red Team (ART) \cite{atomic_red_team_2025} and MITRE ATT\&CK \cite{mitre_attack_2025, al2024mitre} techniques, which enable realistic adversary behavior. The red agent's actions include: 
\begin{itemize}
    \item \textit{Ping sweep:} Identify which hosts are available.
    \item \textit{Port scan:} Identify open ports on a host to understand potential vulnerabilities. 
    \item \textit{Discovery:} Gain more insight into the structure of the network or host to identify additional targets or credentials.
    \item \textit{Lateral movement:} Moving from one host in the network to another host.
    \item \textit{Privilege escalation:} Gaining access to credentials with higher privilege to systems or objects (e.g., files) of interest. 
    \item \textit{Impact:} Achieve the desired objective (e.g., encrypt a system, exfiltrate data).
\end{itemize}
Fig. \ref{fig:network_example} shows an example of different actions of the red agent at play in a particular state of evaluation for a 15-host Cyberwheel network used in this study. The ART agent uses a strategy where the agent gradually starts from low visibility actions, such as portscan, and subsequently moves to lateral movement, and finally makes an impact. 
Fig. \ref{fig:red_agent_yaml} displays an example \texttt{yaml} file defining the baseline red agent's actions and corresponding immediate and recurring rewards. The entry host for the red agent is considered to be random to inject stochasticity in each episode during training and evaluation. The nominal ART agent contains a baseline reward structure based on cyber SME input that provides immediate reward for each action, along with recurring rewards.

In this work, we are interested in evaluating different behavioral traits of a red agent. We first define the qualitative characterizations:
\begin{itemize}
    \item \textit{Aggressive Attacker:} This type of red agent behavior is to reach the stages of privilege escalation, and impact as quickly as possible. The agent intends to gather information about hosts and subnets persistently. As the attacker attempts to inflict maximum damage in a short amount of time, the likelihood of detection is also high.
    \item \textit{Stealthy Attacker:} The red agent in this scenario prioritizes gradual escalation over time, and sustaining a foothold over a longer duration. The agent takes a high degree of caution to avoid detection. 
\end{itemize}


The blue agent pursues a cyber deception strategy and utilizes decoy placements at key locations to impede and neutralize the red agent's progression. The environment can deploy customized decoy hosts on any subnet. The decoy hosts will be inseparable from any normal hosts to the red agent. Blue agent's action space contains three distinct actions: 1) deploy a decoy, 2) remove a decoy, and 3) do nothing. We utilize the baseline DRL-driven blue agent in Cyberwheel, and inject behavioral changes into it. Similar to the Red agent's characterization, we consider the following:

\begin{itemize}
    \item \textit{Proactive Defender version 1:} The blue agent in this case prioritizes taking more frequent actions, having high incentives for decoy placements. The agent intends to maintain a more persistent defensive coverage. 
    \item \textit{Proactive Defender version 2:} The blue agent in this case prioritizes defensive actions under resource constraints. 
\end{itemize}

\subsection{LLM-based Reward Structure Design}

The design of optimal reward structures for red and blue agents in autonomous cyber defense systems presents significant challenges due to the inherent complexity and multi-dimensional nature of cybersecurity operations. As cyber environments evolve with increasingly sophisticated attack vectors, advanced detection methodologies, and expansive network architectures, human designers face substantial cognitive limitations in developing comprehensive reward models that accurately capture the nuanced dynamics of cyber operations. LLMs can aid in guiding the designer with the help of context information of the cyber environment, leveraging their extensive knowledge synthesis capabilities and contextual reasoning abilities. We utilize the Cyberwheel environment to provide the LLM agent (Claude Sonnet 4) with extensive context about the environment and the behavioral tendencies of the agents. The LLM agent helps us translate the qualitative behavioral traits of the red and blue agents to numerical/quantitative measures of the actions by suggesting different rewards based on the characteristics of the agents. Claude Sonnet 4.0 leads on the SWE-bench Verified benchmark, which evaluates an LLM's ability to solve real-world software issues using code samples verified to be non-problematic by our human annotators \cite{openai2024swebench}. 
Claude Sonnet 4.0 also excels at other benchmarks like Terminal-Bench \cite{tbench_2025}, GPQA Diamond \cite{rein2024gpqa}, TAU-bench \cite{yao2024tau}, Multilingual Q\&A (MMMLU) \cite{hendryckstest2021}, and Visual reasoning
(MMMU) \cite{yue2023mmmu}. We provided the initial contextual inputs about the cyber simulation environment to the LLM agent before the behavioral prompts. We present representative prompts that were used to probe the LLM agent to generate reward structures based on the behavioral characteristics of the agents as follows.

\begin{tcolorbox}[myblurbB, title=\textit{Prompt for attack (red) agent to be aggressive:}]
Consider that you are an expert in cyber operations. How can the attacker (red) agent be made more aggressive? Where the red agent follows the Atomic Red Team (ART) and MITRE ATT\&CK strategy. Using the baseline yaml file given with this prompt, provide modifications for an aggressive red agent.
\end{tcolorbox}

\begin{tcolorbox}[myblurbA, title= \textit{Prompt for defense (blue) agent to be proactive:}]
Consider you are an expert in cyber operations. How can the defender (blue) agent be made more proactive, where the blue agent’s action spaces are: doing nothing, decoy placement (decoy0), and removing decoy? Using the baseline yaml file given with this prompt, provide modifications for a proactive blue agent. 
\end{tcolorbox}

\subsection{Optimal Cyber Defense Policy Design}
\begin{algorithm}[t!]
\caption{LLM-assisted Proximal Policy Optimization (PPO) with Clipped Surrogate Objective}
\begin{algorithmic}[1]
    \STATE \textbf{Input:} Cyberwheel environment $\mathcal{E}$, policy $\pi_\theta$, value function $V_\phi$, clip parameter $\epsilon$, learning rates $\alpha_\theta, \alpha_\phi$
    \STATE \textbf{Reward Structure Generation via LLM:}
        \STATE  Query an LLM (e.g., Claude Sonnet 4) with red and blue agent behavioral traits to obtain reward structures
        \begin{align*}
        \mathcal{R}_{\text{LLM}} = \mathcal{L}_{\text{LLM}}(\text{task }&\text{prompt, context,}\\
         &\text{behavioral constraints})
        \end{align*}
        
        \STATE Optionally refine $\mathcal{R}_{\text{LLM}}$ using expert feedback and environment constraints.

    \STATE \textbf{Policy Optimization:}
        \FOR{each training iteration}
            \STATE Collect trajectories $\tau = \{(s_t, a_t, r_t, s_{t+1})\}$ using current policy $\pi_{\theta_{old}}$ where $r_t = f(\mathcal{R}_{\text{LLM}})$.
            \STATE Compute advantage estimates $A_t$ (e.g., using GAE)
            \FOR{each epoch and minibatch of data}
                \STATE Compute the policy ratio:
                \[
                r_t(\theta) = \frac{\pi_\theta(a_t|s_t)}{\pi_{\theta_{old}}(a_t|s_t)}
                \]
                \STATE Evaluate the clipped surrogate objective $L_t^{CLIP}(\theta)$ as in \eqref{eqn:surrogate_objective}.
                \STATE Update policy parameters:
                \[
                \theta_{k+1} \leftarrow \theta_k + \alpha_\theta \nabla_\theta L_t^{\text{CLIP}}(\theta_k)
                \]
                \STATE Update value function parameters:
                \[
                \phi_{k+1} \leftarrow \phi_k - \alpha_\phi \nabla_\phi \mathbb{E}_t[(V_\phi(s_t) - R_t)^2]
                \]
            \ENDFOR
            \STATE Set $\pi_{\theta_{old}} \leftarrow \pi_\theta$
        \ENDFOR
\end{algorithmic}
\end{algorithm}

\begin{table*}[!t]
\centering
\begin{tabular}{ll}
\toprule
\hline
\textbf{Parameter} & \textbf{Description / Value} \\
\hline
\midrule
$\alpha_\theta, \alpha_\phi$ & $2.5\times10^{-4}$ \, (Learning rates for policy and value networks) \\
\texttt{anneal\_lr} & \texttt{true} \, (Anneal learning rate for policy/value nets) \\
$\gamma$ & $0.99$ \, (Discount factor) \\
$\lambda$ & $0.95$ \, (GAE parameter) \\
\texttt{num\_minibatches} & $4$ \, (Number of minibatches) \\
 \texttt{num\_epochs} & $4$ \, (Number of update epochs) \\
$\epsilon$ & $0.2$ \, (Surrogate clipping coefficient) \\
$\beta$ & $0.01$ \, (Entropy coefficient) \\
$c_v$ & $0.5$ \, (Value function coefficient) \\
$\|\nabla\|_{\max}$ & $0.5$ \, (Maximum gradient norm) \\
\hline
\bottomrule
\end{tabular}
\caption{DRL hyperparameters.}
\label{tab:rl_params_symbols}
\end{table*}

\begin{figure*}[!t]
    \centering
    \begin{subfigure}[b]{0.32\textwidth}
        \centering
        \includegraphics[width=\textwidth]{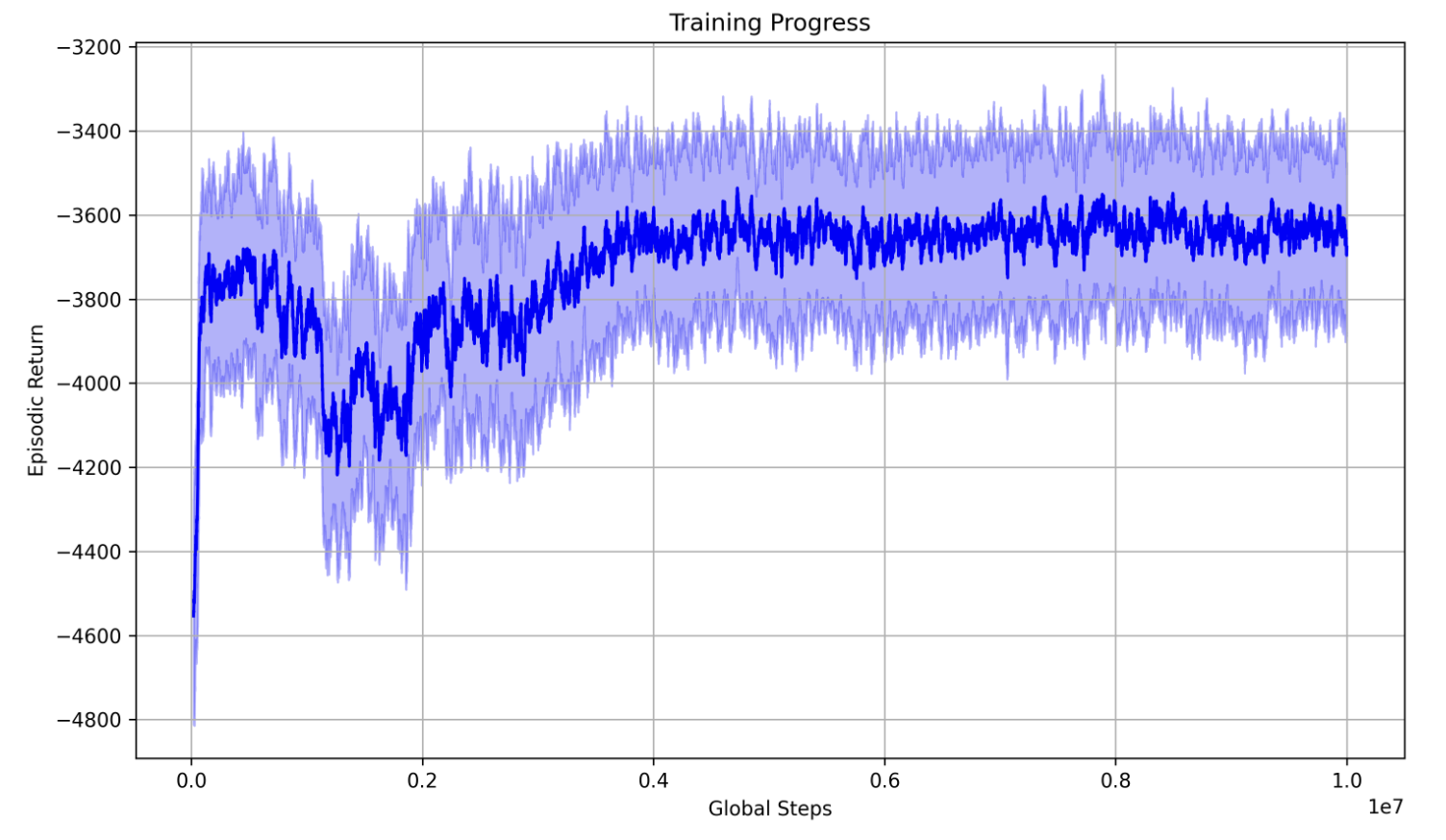}
        \caption{Baseline blue agent}
        \label{fig:sub1}
    \end{subfigure}
    \hfill
    \begin{subfigure}[b]{0.32\textwidth}
        \centering
        \includegraphics[width=\textwidth]{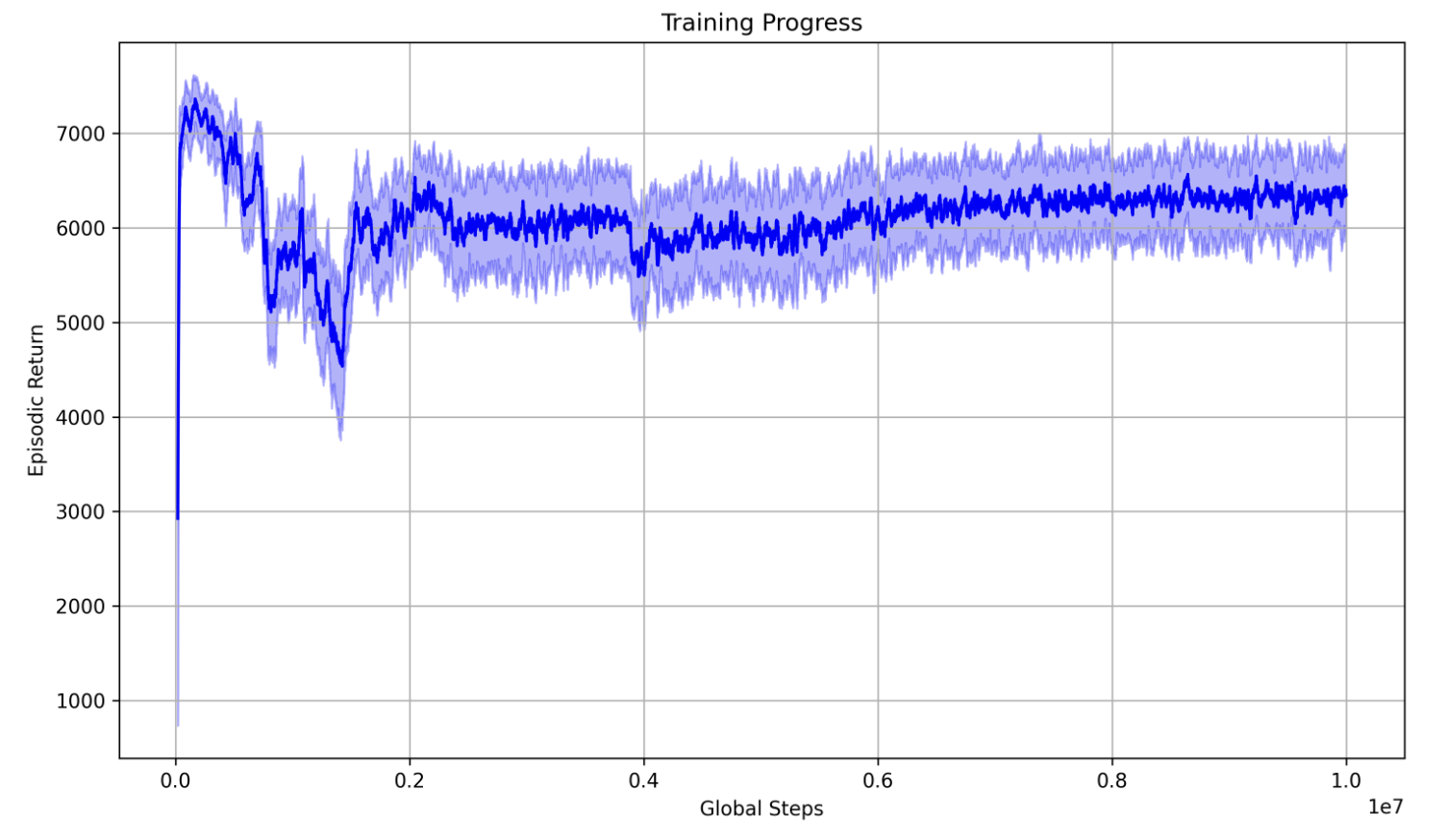}
        \caption{Blue agent proactive-v1}
        \label{fig:sub2}
    \end{subfigure}
    \hfill
    \begin{subfigure}[b]{0.32\textwidth}
        \centering
        \includegraphics[width=\textwidth]{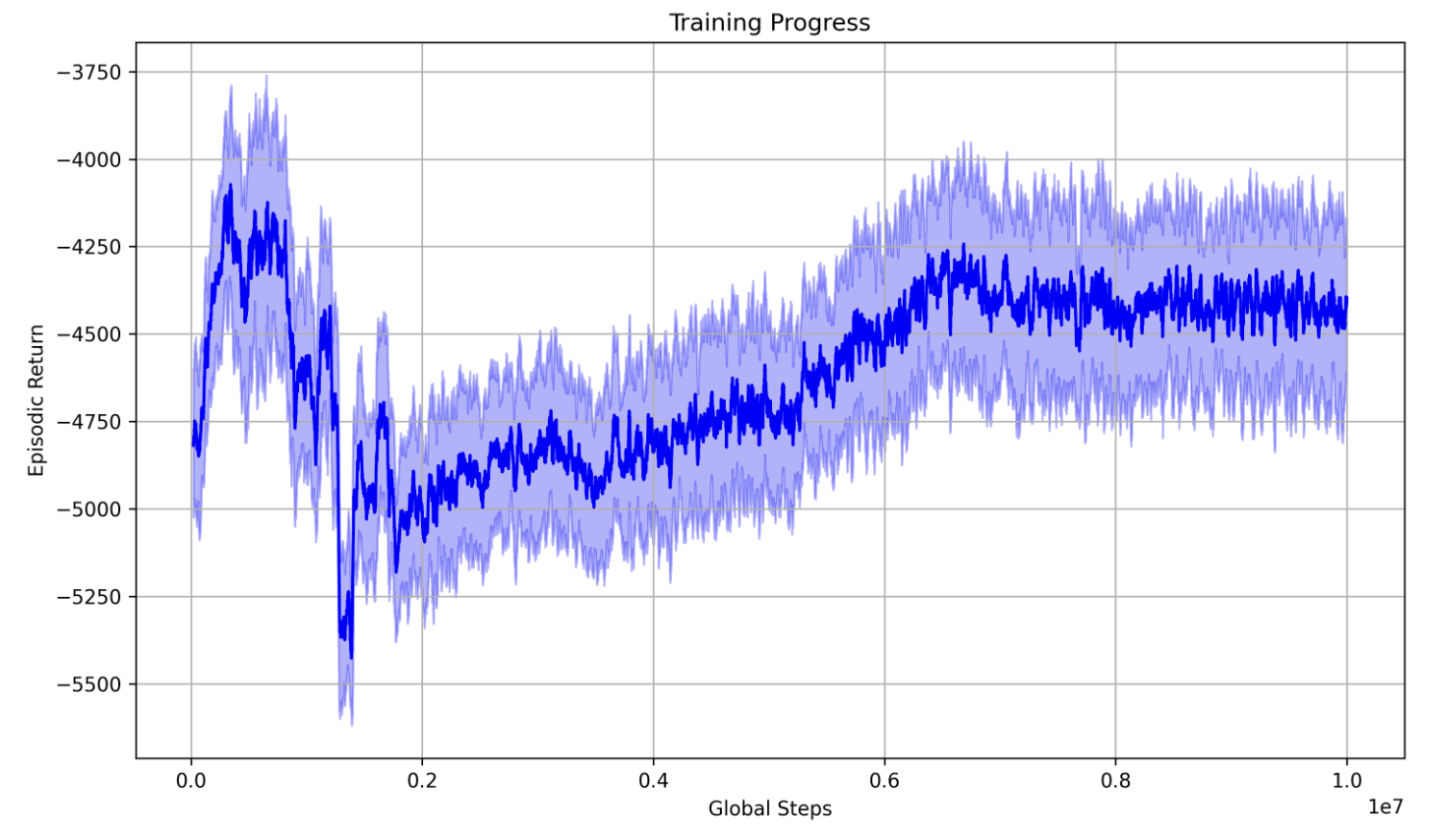}
        \caption{Blue agent proactive-v2}
        \label{fig:sub3}
    \end{subfigure}
    \caption{Training progress of blue agent for different scenarios against a baseline red agent.}
    \label{fig:training}
\end{figure*}

\noindent
Details of the LLM-guided PPO-based DRL cyber defense training mechanism is provided in Algorithm 1. The blue agent's action is sampled from a categorical distribution. The policy network is parametrized by $\theta$, and computed as:
\begin{align}
    \pi_\theta(a|s) = \text{Categorical}(z_\theta(s))_a,
\end{align}
where generally this categorical distribution is computed via softmax functions, and $z_\theta(s)$ computes the logits. Subsequently, at each step, the action is sampled from the policy distribution $a_t \sim \pi(.|s_t)$. At each time step during training, the reward is constructed using the blue agent's action-based rewards, along with the red agent's, creating a composite reward. The recurring costs are also considered over a particular episode. Therefore, at time step $t$, we have:
\begin{align}
    r_{b,t}, r_{r,t} ,r_{b,t}^{reccur}, r_{r,t}^{reccur}\leftarrow \mathcal{R}_{LLM},\\
    r_t = r_{b,t} + r_{r,t} + \sum_{k=0:t} (r_{b,t}^{reccur}+ r_{r,t}^{reccur}).
\end{align}

When the red agent compromises legitimate assets, a negative reward for the defender and a persistent penalty are applied. Whereas, if the red agent successfully attacks a decoy infrastructure, then high incentives (positive numbers) are considered. The environment is equipped with a detector that creates alerts if the red agent's action target is a decoy. The observation
space (here similar interpretation to observable states $s_t$) is constructed by tabulating if the detector produced an alert for the non-decoy host at that time step in a Boolean form, and a history of that non-decoy host if the detector ever produced an alert. We employ an actor-critic method-based DRL agent, using proximal policy optimization (PPO) \cite{schulman2017proximal}. Actor-Critic methods integrate elements of both policy iteration and value iteration. Critic performs policy evaluation by minimizing the critic loss and computing gradients that are passed on to the actor. 
The actor is responsible for selecting the optimal action based on the current state. The parameters of this network are adjusted using the gradients received from the critic. The actor aims to find the optimal parameter set $\theta^*$ that maximizes the expected cumulative reward~\cite{sutton2018reinforcement}:
\begin{equation}
\label{eqn:policy_iteration}
J(\pi_{\theta}) = E_{\pi_{\theta}}\left[\sum_{t=1}^T r(s_t, a_t)\right]
\end{equation}
where $T$ denotes the total number of decision steps. The update rule for the actor’s parameters is given by $\theta = \theta + \alpha \nabla J(\theta)$
where $\alpha$ is the learning rate. The policy gradient $\nabla J(\theta)$ is computed as follows~\cite{sutton2018reinforcement}:
\begin{equation} \label{eqn:gradient_desc} \begin{split} \nabla J(\pi_{\theta})& =\nabla E_{\pi_{\theta}}[\sum_{t=1}^Tr(s_t,a_t)] \\ & = \nabla E_{\pi_{\theta}}(\sum_{t=1}^T\nabla_{\theta}log \pi_{\theta}(a_t|s_t)r(s_t,a_t)). \end{split} \end{equation}
PPO mitigates large policy updates by constraining the new policy within a specified range of the old one. It introduces a clipping mechanism to restrict the probability ratio to the interval $[1-e, 1+e]$, where $e$ is a hyperparameter controlling the extent of deviation. This helps improve stability and robustness, especially in the presence of sensor noise or model errors. PPO maximizes the following clipped surrogate objective:
    \begin{equation}\label{eqn:surrogate_objective}
    \begin{split}
        L_t^{CLIP}(\theta) = \mathbb{E}[\min \{
        w_t(\theta) A_t, \\
        \textrm{clip}(w_t(\theta), 1 - e, 1 + e) A_t\}]
    \end{split}
    \end{equation}
    where $w_t(\theta) = \frac{\pi_{\theta}(a_t|s_t)}{\pi_{old}(a_t|s_t)}$ denotes the probability ratio between the new and old policies, and $A_t$ is the advantage estimate at time step $t$.

\begin{table}[t!]
\centering
\begin{tabular}{|l|c|c|}
\hline
\textbf{Action Name} & \textbf{\begin{tabular}{c}Immediate\\Reward\end{tabular}} & \textbf{\begin{tabular}{c}Reccuring\\Reward\end{tabular}} \\
\hline
nothing & 0.0 & 0.0 \\
\hline
decoy0 & -20.0 & 0.0 \\
\hline
remove\_decoy & -1.0 & 0.0 \\
\hline
\end{tabular}
\caption{Baseline blue agent actions and rewards.}
\label{tab:baseline_blue_agent_rewards}
\centering
\vspace{0.3 cm}
\begin{tabular}{|l|c|c|}
\hline
\textbf{Action Name} & \textbf{\begin{tabular}{c}Immediate\\Reward\end{tabular}} & \textbf{\begin{tabular}{c}Reccuring\\Reward\end{tabular}} \\
\hline
nothing & -5.0 & -1.0 \\
\hline
decoy0 & 20.0 & 2.0 \\
\hline
remove\_decoy & -10.0 & -2.0 \\
\hline
\end{tabular}
\caption{Proactive-v1 blue agent actions and rewards.}
\label{tab:proactive_blue_agent_rewardsv1}
\centering
\vspace{0.3 cm}
\begin{tabular}{|l|c|c|}
\hline
\textbf{Action Name} & \textbf{\begin{tabular}{c}Immediate\\Reward\end{tabular}} & \textbf{\begin{tabular}{c}Recurring\\Reward\end{tabular}} \\
\hline
nothing & -5.0 & -1.0 \\
\hline
decoy0 & -5.0 & -0.5 \\
\hline
remove\_decoy & -10.0 & 0.0 \\
\hline
\end{tabular}
\caption{Proactive-v2 blue agent actions and rewards.}
\label{tab:proactive_blue_agent_rewardsv2}
\end{table}

\begin{table}[h]
\centering
\begin{tabular}{|l|c|c|}
\hline
\textbf{Action Name} & \textbf{\begin{tabular}{c}Immediate\\Reward\end{tabular}} & \textbf{\begin{tabular}{c}Recurring\\Reward\end{tabular}} \\
\hline
pingsweep & 1 & 0 \\
\hline
portscan & 2 & 0 \\
\hline
discovery & 5 & 0 \\
\hline
lateral-movement & 10 & 0 \\
\hline
privilege-escalation & 20 & 0 \\
\hline
impact & 50 & 0 \\
\hline
\end{tabular}
\caption{Baseline red agent actions and rewards.}
\label{tab:baseline_red_agent_rewards}
\vspace{0.3 cm}
\centering
\begin{tabular}{|l|c|c|}
\hline
\textbf{Action Name} & \textbf{\begin{tabular}{c}Immediate\\Reward\end{tabular}} & \textbf{\begin{tabular}{c}Recurring\\Reward\end{tabular}} \\
\hline
pingsweep & 5 & 2 \\
\hline
portscan & 10 & 3 \\
\hline
discovery & 20 & 5 \\
\hline
lateral-movement & 40 & 15 \\
\hline
privilege-escalation & 75 & 25 \\
\hline
impact & 150 & 50 \\
\hline
\end{tabular}
\caption{Aggressive red agent actions and rewards.}
\label{tab:aggressive_red_agent_rewards}
\centering
\vspace{0.3 cm}
\begin{tabular}{|l|c|c|}
\hline
\textbf{Action Name} & \textbf{\begin{tabular}{c}Immediate\\Reward\end{tabular}} & \textbf{\begin{tabular}{c}Recurring\\Reward\end{tabular}} \\
\hline
pingsweep & 0.5 & 3 \\
\hline
portscan & 1 & 5 \\
\hline
discovery & 3 & 8 \\
\hline
lateral-movement & 5 & 20 \\
\hline
privilege-escalation & 8 & 25 \\
\hline
impact & 15 & 50 \\
\hline
\end{tabular}
\caption{Stealthy red agent actions and rewards.}
\label{tab:stealthy_red_agent_rewards}
\end{table}
\section{Experimental Results and Discussion}
\label{sec:experiments}

We conducted a set of experiments to evaluate how the defense agent policies are trained based on different behavioral characterizations of attack and defense agents using the 15-host network, as shown in Fig. \ref{fig:network_example}. The blue agent's DRL training was performed using the hyperparameters provided in Table \ref{tab:rl_params_symbols}. The Cyberwheel environment contains SME-guided baseline reward structures of the blue agent as provided in Table \ref{tab:baseline_blue_agent_rewards}. With the help of Claude-Sonnet-4, we queried the LLM agent to make the blue RL agent more proactive, and the resultant reward structure is given in Table \ref{tab:proactive_blue_agent_rewardsv1}. We note this blue agent as \textit{proactive-v1} where the LLM agent has provided a high incentive for placing the decoy. We then refined our queries and asked for a reward structure where the blue agent will still encounter some amount of cost for placing the decoy, although due to the proactive nature, the penalty is much less than baseline, which is denoted by \textit{proactive-v2} policy. The defender is also substantially penalized for doing nothing for both \textit{proactive-v1} and \textit{proactive-v2} policies. Subsequently, we also queried the LLM agent to impart variations in characterizations in how the ART red agent's actions are perceived. Tables \ref{tab:baseline_red_agent_rewards}, \ref{tab:aggressive_red_agent_rewards}, and \ref{tab:stealthy_red_agent_rewards} present the baseline, and LLM-informed \textit{aggressive} and \textit{stealthy} personas of the attacker. We can infer from these tables that with more aggressive behavior, the immediate rewards for lateral movement, privilege escalation, and impacts make these actions more concerning for the blue agent. In contrast, for the stealthy persona, the red agent's action rewards are decreased by the LLM. The stealthy red agent's actions have lower immediate rewards; however, the recurring rewards are quite high, indicating the characterization of gradual escalation. 
We conducted a total of $9$ experiments to account for all combinations of the blue agent as baseline, \textit{proactive-v1}, and \textit{proactive-v2} with the red agent as baseline, \textit{stealthy} and \textit{aggressive}. Fig. \ref{fig:training} shows the training performance with episodic rewards for different baseline and proactive policies against a baseline red agent. Once the policies were trained, we ran $50$ evaluation episodes with $100$ steps for each episode. 

\begin{figure}[t!]
    \centering
    \includegraphics[width=\linewidth]{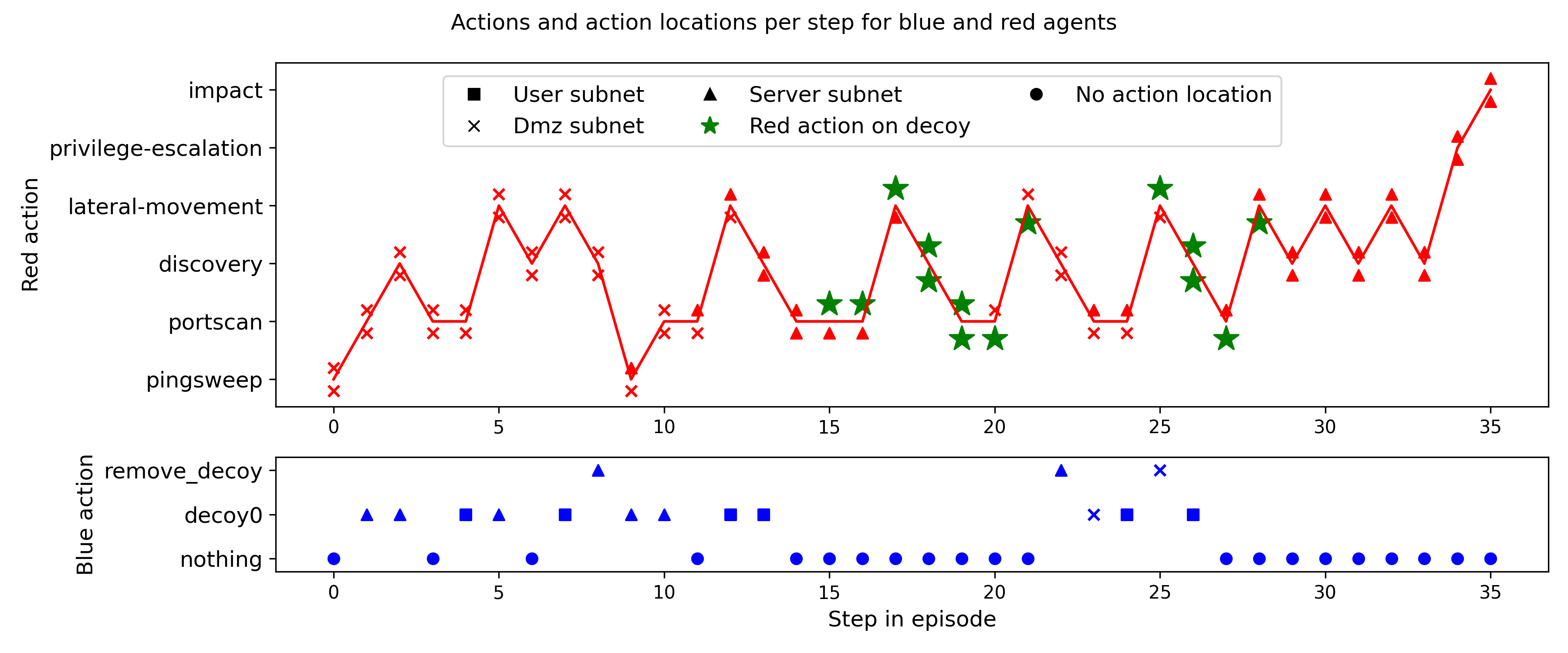}
    \caption{A single episode of red agent's action steps (top) and blue agent actions. For each step in the red agent's trajectory the lower symbol corresponds to the subnet where the action originates (its source) and the upper symbol corresponds to the subnet where the action takes place (its destination).}
    \label{fig:trajectory}
\end{figure}

As an example of a single episode we show the actions of the \emph{stealthy} red agent and \emph{proactive-v1} blue agent over time in Fig. \ref{fig:trajectory}.
This represents typical behavior where the red agent moves between reconaissance steps (pingsweep, portscan), discovery, and lateral movement until it finally can escalate privileges and make impact. Notice that the red agent makes some actions on decoys (green stars) that are deployed by the blue agent throughout the episode.
\begin{figure}[t!]
    \centering
    \includegraphics[width=\linewidth]{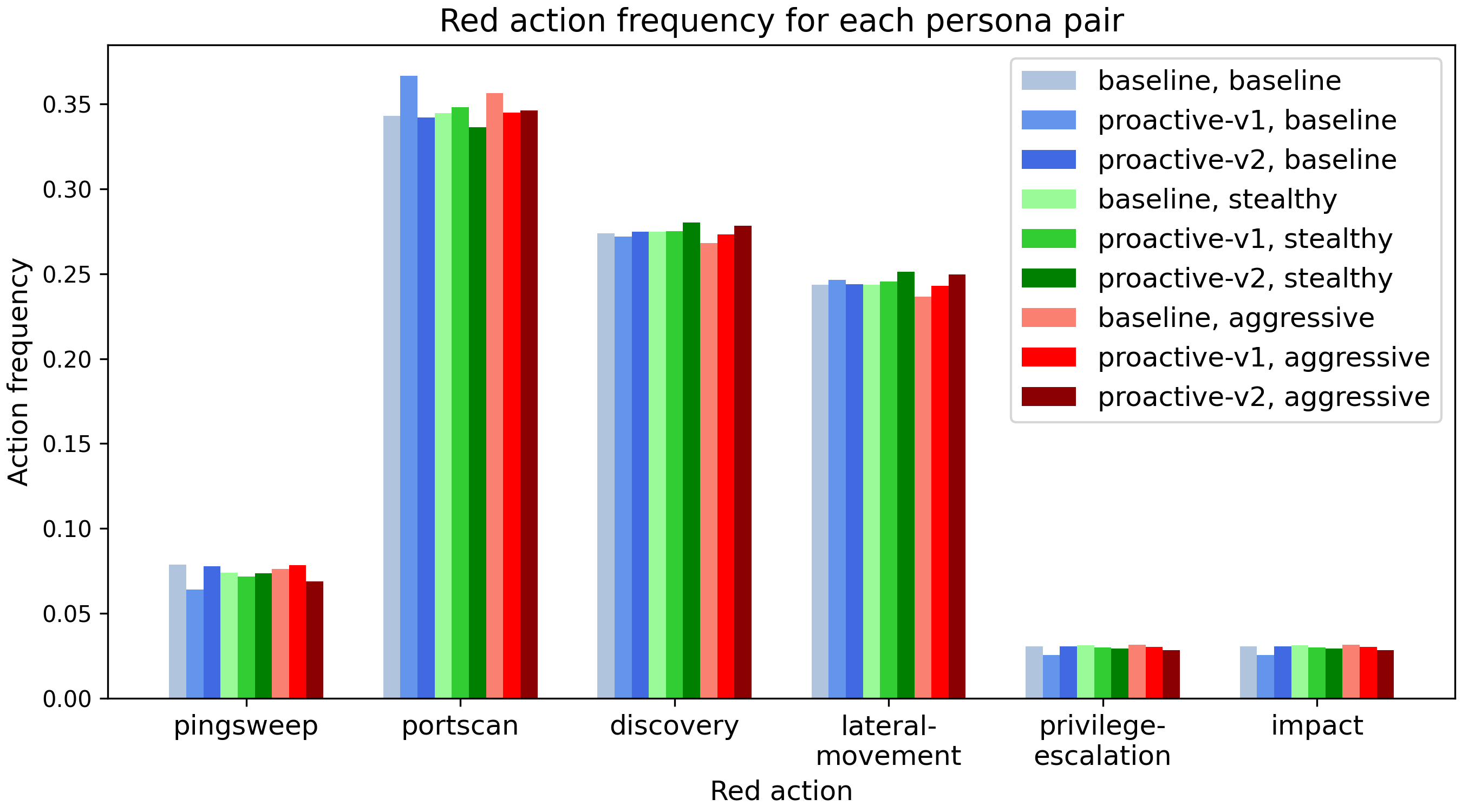}
    \caption{Frequency of red agent actions for each pair of (blue, red) personas. Since the red agent is not a learned agent, we do not see significant changes in red agent behavior. The change is in the reward structures and how the blue learning agent perceives the red actions.}
    \label{fig:red_actions}
\end{figure}
\begin{figure}[t!]
    \centering
    \includegraphics[width = \linewidth]{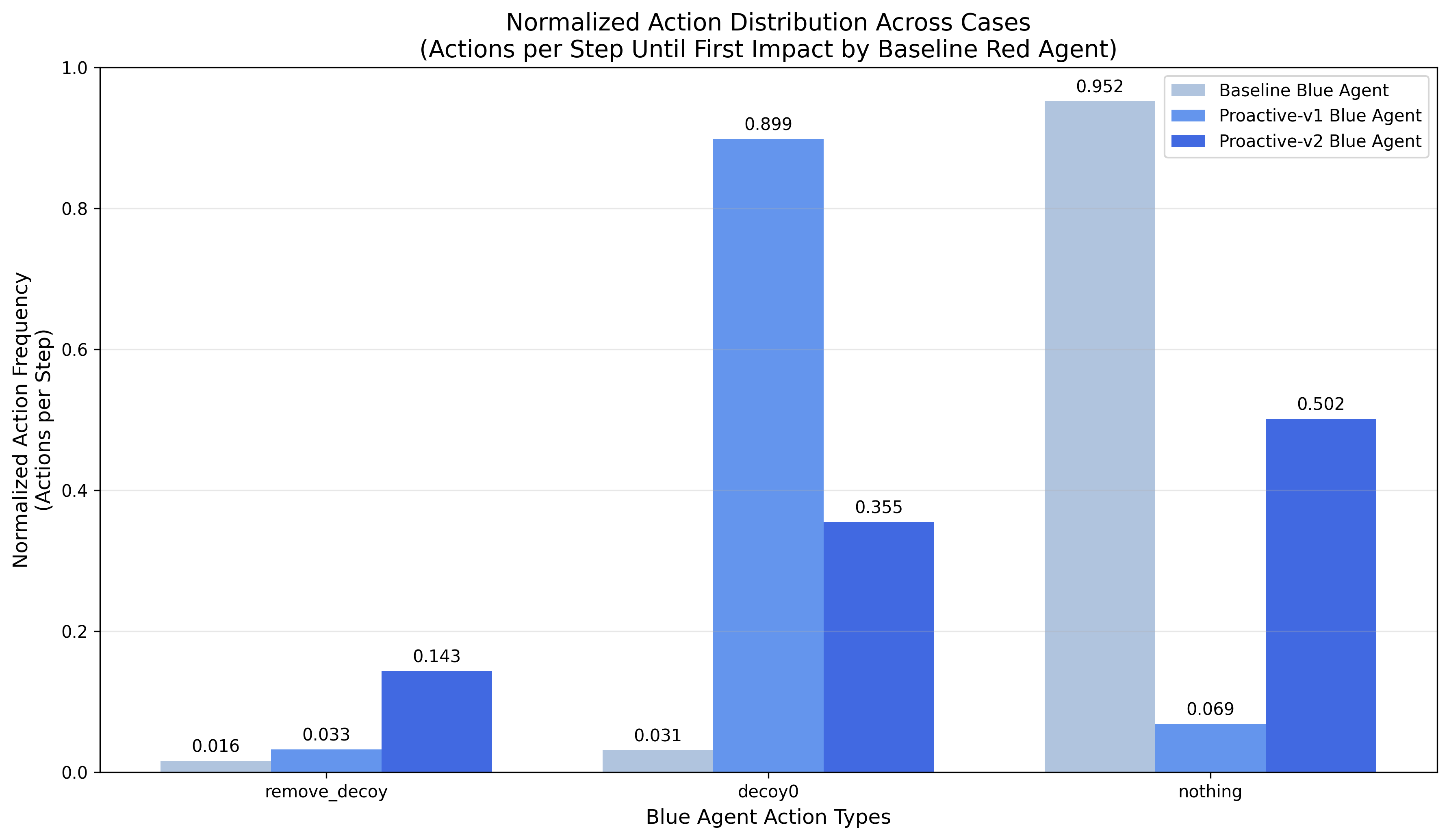}
    \includegraphics[width = \linewidth]{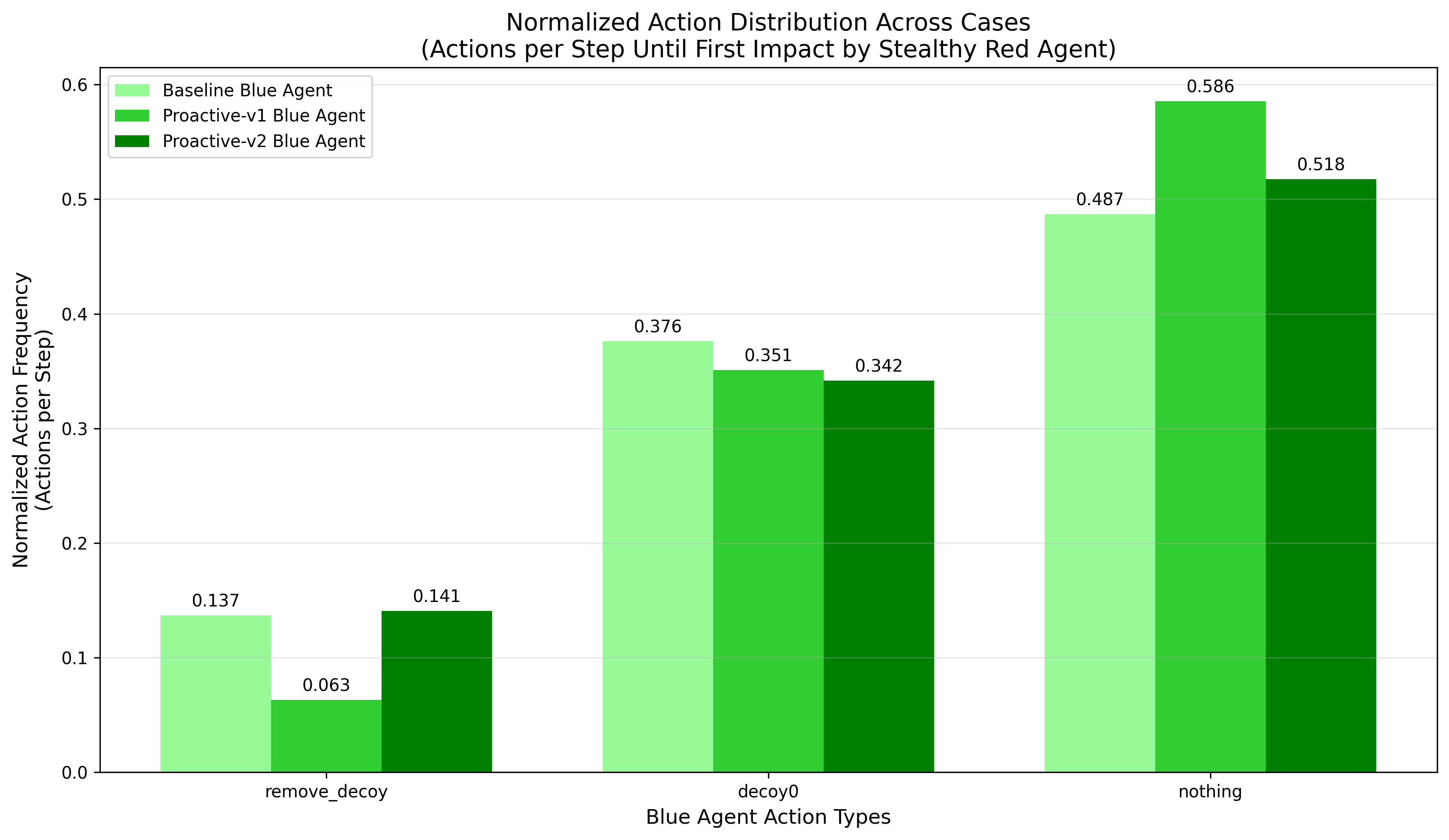}
    \includegraphics[width = \linewidth]{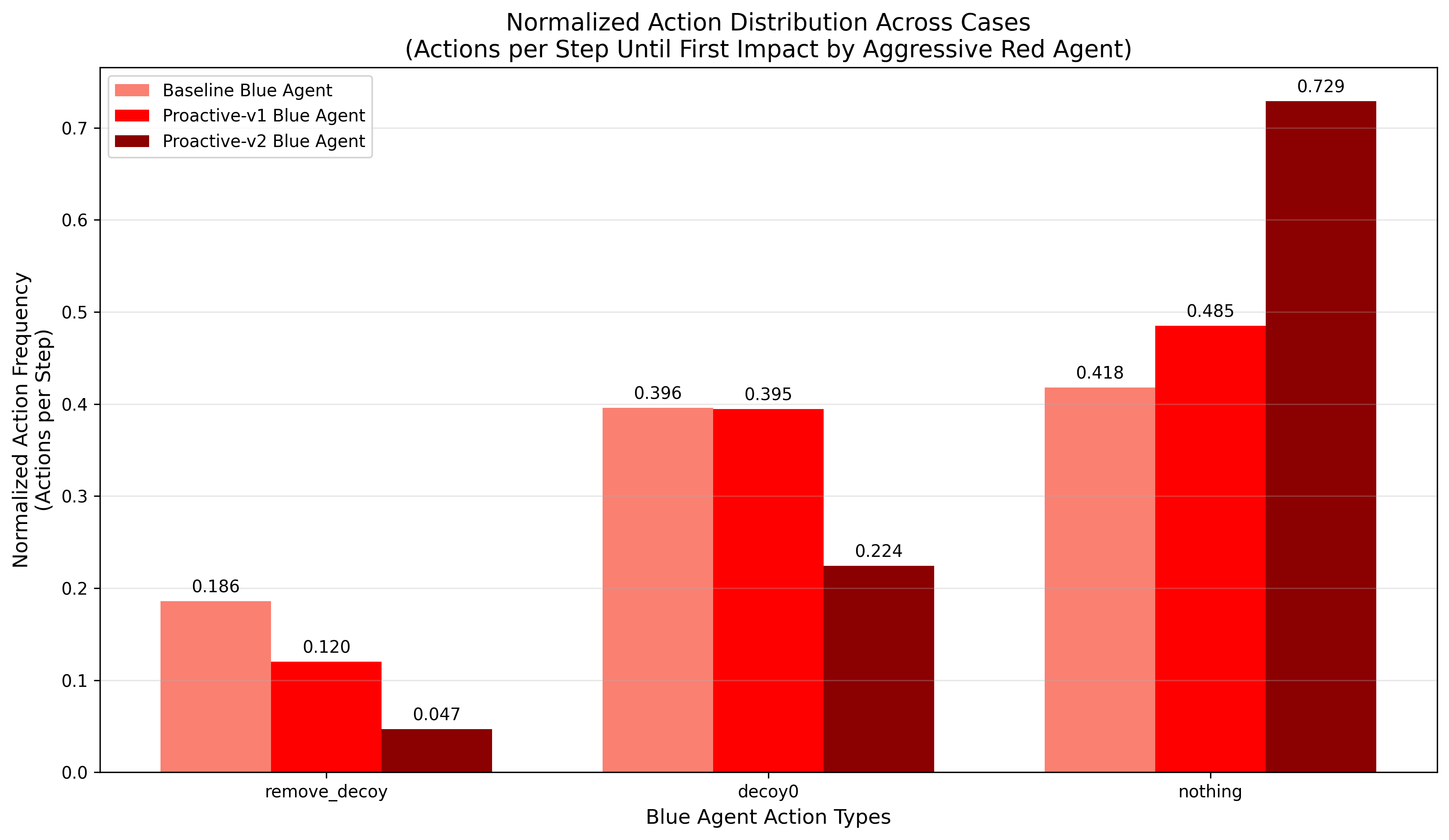}
    \caption{Actions of baseline, proactive-v1, and proactive-v2 blue agents (both counts and normalized frequencies) against baseline (top), stealthy (middle), and aggressive (bottom) red agents.}
    \label{fig:freq_blue}
\end{figure}
\begin{figure*}[t!]
    \centering
    \includegraphics[width=1\textwidth]{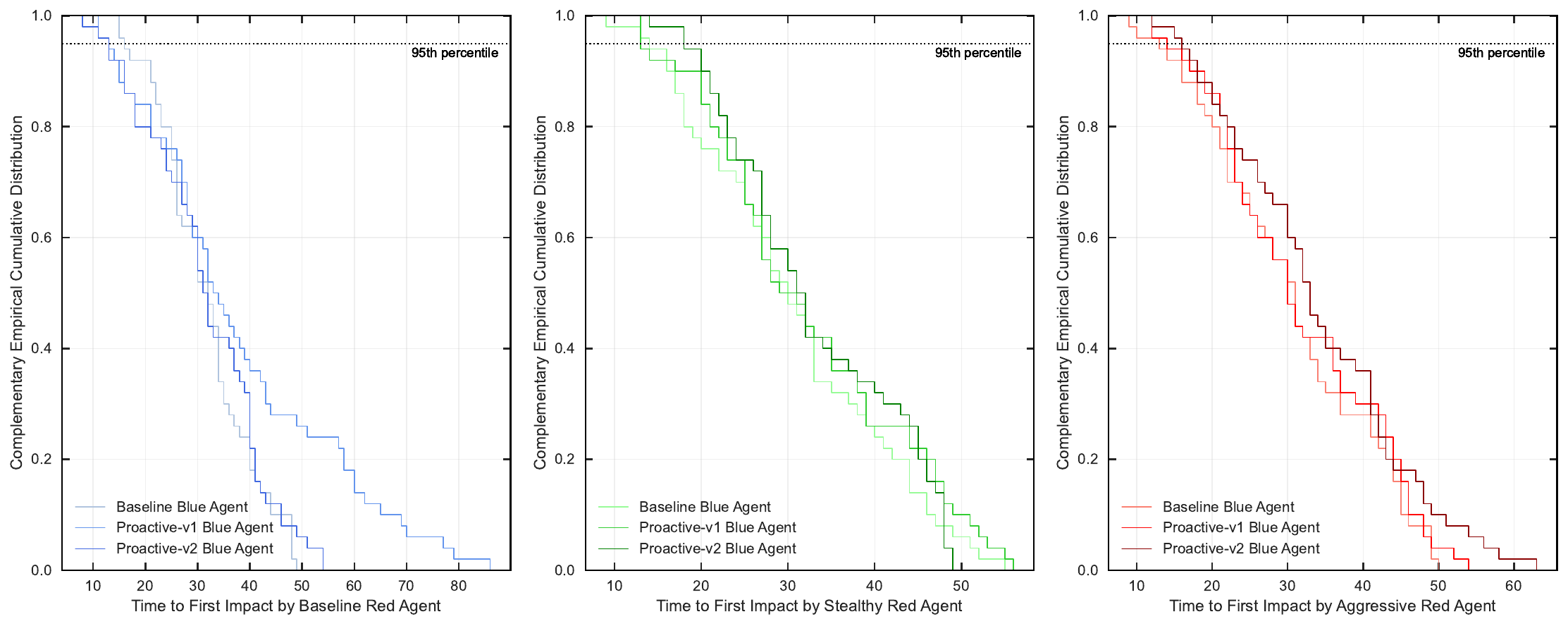}
    \caption{Complementary empirical cumulative distributions of time to first impact by heuristic-based red agents against DRL-based blue agents.}
    \label{fig:fig_ccdf}
\end{figure*}
In our experiments, the blue agent is the only learning agent, while the red agent follows a heuristic. Therefore, it is important to observe that the red agent is not changing its behavior. In Fig. \ref{fig:red_actions} we show the frequency of each red action across all (blue, red) persona combinations to see that the red agent does not do each action more or less frequently across the different personas. Instead, what is changed is how the blue agent perceives the actions of the red agent. The blue agent may not consider the stealthy red agent's actions problematic because of the low reward value and, therefore may deploy fewer decoys and do nothing more often. In contrast, the blue agent may be very worried about the aggressive red agent's actions because of the high red reward value, which could cause the blue agent to deploy more decoys. 

To evaluate a blue agent's behavior resulting from LLM-based reward functions, 
we explored two questions:

\begin{enumerate}
    \item \textit{What are the action distributions for blue agents in response to different red agent personas?}
    \item \textit{How successful are the blue agents against different red agent personas?}
\end{enumerate}

\noindent
To address the first question, action distributions of the baseline, proactive-v1, and proactive-v2 blue agents (normalized frequencies with respect to the number of steps before red agent's first impact) against baseline (top), stealthy (middle), and aggressive (bottom) red agents were plotted and are displayed in Fig. \ref{fig:freq_blue}. Against the baseline red agent, the proactive-v1 agent places substantially more decoys than the baseline and the proactive-v2 blue agent. The proactive-v1 agent takes more decoy-placing actions against the aggressive red agent ($39.5\%$)  than the stealthy red agent ($35.1\%$), showing a more proactive trait against the aggressive attacker. On the other hand, for the proactive-v2 agent, the proportion of decoys against the stealthy red agent ($34.2\%$) is much higher than the aggressive red agent ($22.4\%$). 

To address the second question we subsequently evaluate a criterion of the success of the defender in terms of the number of steps required for the red agent to perform the first impact. Fig. \ref{fig:fig_ccdf} displays complementary empirical cumulative distributions characterizing the minimum time to first impact, $X$ for red agents (i.e., $1-F_X(x)=P(X>x)$), with data from $50$ evaluation episodes with $100$ steps per episode, operating against various DRL-trained blue agent policies. The $95^{th}$ percentile value in these plots (i.e., $95\%$ probability that the red agent's first time to impact will exceed a particular value) serves as a robust metric to measure the lower bound of time steps (i.e., worst case scenario) required by different red agents to successfully execute their first impact action across the defense policy spectrum. Against the baseline red agent, the baseline blue agent demonstrates the best performance with $95^{th}$ percentile value for time to first impact as $15$ time steps compared to proactive-v1 and proactive-v2 blue agent values of $11$ time steps, delaying the steps-till-impact measure against the attacker. Against the stealthy red agent, the proactive-v2 blue agent demonstrates the best performance with $95^{th}$ percentile value for time to first impact as $18$ time steps compared to proactive-v1 and baseline blue agent values of $13$ time steps. Finally, against the aggressive red agent, the proactive-v2 blue agent ($15$ time steps) again outperforms proactive-v1 ($12$ time steps) and baseline ($10$ time steps) blue agents. These simulation results suggest that an effective blue agent policy against diverse red agent personas is to employ a mixed strategy where a baseline blue agent operates against a baseline red agent, and switches to a proactive-v2 blue agent when encountered with a stealthy or aggressive red agent.

\section{Conclusion}
\label{sec:conclusion}
Using a context-aware LLM for reward design, with heterogeneous attack and defense agents, within a DRL-driven autonomous cyber defense setting is a promising approach. Our simulation experiments suggest that LLM-informed proactive DRL defense agents lead to greater delays in \textit{impact} action across stealthy and aggressive types of heuristic-based red agents. Also, an effective blue agent policy against diverse red agent personas is to employ a mixed strategy where the blue agent can switch personas depending on the red agent persona that it may encounter. There are several future directions to pursue further, including enabling the red agent to be a learning agent in addition to the blue agent, including multiple red and blue agents, scaling up state space of the cyber host network, and pursuing an LLM-in-the-loop approach for reward design. The latter direction would involve having an LLM suggest a reward structure, train the agent, and allow the LLM to iteratively refine the reward structure to improve the blue agent's performance. We believe that including an LLM in the process, either in initial reward design as performed in this work or in the loop to refine rewards as proposed, can help make DRL-based autonomous cyber defense agents more robust to changing adversary tactics and make their actions more explainable.

\section{Acknowledgments}
This research was supported by the Generative AI (GenAI) investment at Pacific Northwest National Laboratory (PNNL) under the Laboratory Directed Research and Development (LDRD) program. PNNL is multiprogram national laboratory operated by Battelle for the U.S. Department of Energy under contract DE-AC05-76RL01830. The authors thank Sean Oesch and Amul Chaulagain at Oak Ridge National Laboratory for helpful discussions about Cyberwheel simulation environment.

\bibliography{aaai2026}

\end{document}